\pdfoutput=1

\documentclass[11pt]{article}

\usepackage{emnlp2021}
\usepackage{graphicx}

\usepackage{times}
\usepackage{latexsym}

\usepackage[T1]{fontenc}

\usepackage[utf8]{inputenc}

\usepackage{microtype}

\begin{document}

\title{CipherSniffer: Classifying Cipher Types}
\author{Brendan Artley, Greg Mehdiyev\\
  Simon Fraser University \\
  Burnaby, BC, Canada \\
  \texttt{\{brendan.artley,gma60\}@sfu.ca} \\ \\ 
  }
\date{\today}
\maketitle

\begin{abstract}
Ciphers are a powerful tool for encrypting communication. There are many different cipher types, which makes it computationally expensive to solve a cipher using brute force. In this paper, we frame the decryption task as a classification problem. We first create a dataset of transpositions, substitutions, text reversals, word reversals, sentence shifts, and unencrypted text. Then, we evaluate the performance of various tokenizer-model combinations on this task.
\end{abstract}

\section{Introduction}

Identifying ciphers is a challenging problem. Human code-breakers typically do this by looking for patterns in the encrypted text. Common strategies include analyzing common letter pairings, counting character frequencies, and trying to solve the small words first. These strategies work well for humans, but we want to see if language models can pick up on these strategies as well.

We evaluated different tokenization methods and model types on this problem. We were interested in comparing word-level tokenizers and subword-level tokenizers on this task. This problem is challenging because some cipher methods shuffle the order of characters arbitrarily, resulting in a large amount of out-of-bounds vocabulary. It was interesting to see how the models adapted to pick up on cipher hints rather than the semantic meaning from the text.

\section{Related Work}

Deciphering classic ciphers with language models is an area with existing research. In \citet{berg-kirkpatrick-klein-2013-decipherment} the authors implement a trigram HMM model using beam search to solve substitution ciphers. They show how using HMMs with millions of random restarts results in significant increases in decipherment. This research was enhanced by \citet{kambhatla-etal-2018-decipherment}, where the authors showed how to reduce the search space of the beam search algorithm by scoring the entire plaintext at each decipherment step. These papers focus on models that directly solve ciphers, but could be improved by the classification models in our paper. 

We also acknowledge the work done by \citet{krishna-etal-2018-classify} in classifying ciphers using SVMs, HMMs, and CNNs. They find that an HMM architecture is the most accurate classifier. Also, the author shows how models get more accurate as the length of text increases. They experiment with cipher lengths between 10 and 10000 characters whereas we experiment with cipher lengths between 7 and 443 characters.

Finally, we drew inspiration for the cipher types from \citet{jumbled-letters} and \citet{robust-word-recognition}. In \citet{jumbled-letters}, the authors show that people find it difficult to identify words when the start and end characters are jumbled rather than when the internal characters are jumbled. \citet{robust-word-recognition} takes this further by implementing a semi-character level RNN that is more robust on a spelling correction task. They do this by embedding the starting character, ending character, and internal character counts in three separate vectors. These two papers motivated us to create the word reversal, text reversal, and character shift ciphers.

\section{Approach}

We framed our task as a 6-label classification problem, with 5 cipher classes and 1 unencrypted text class. Then, we experimented with various model and tokenizer combinations and evaluated each model based on its performance on a hidden test set.

\subsection{Ciphers}

The five ciphers that we used were substitution cipher, transposition cipher, word reversal, text reversal, and character shift. Except for text reversal, each encipherment method preserves the position of the whitespace in the text. We chose these ciphers as they each modify the text differently.

When we applied each cipher to the dataset, we iteratively looped through the cipher types. This ensured that the dataset had an equal class distribution and made training and evaluation easier. The data is saved in a txt file format, with the first character of each line corresponding to its class label. The label encoding that we used is shown in table~\ref{tab:EncodedLabels}.

\begin{table}
\centering
\begin{tabular}{lc}
\hline
\textbf{Label} & \textbf{Encoded Label}\\
\hline
\verb|Substitution| & {0} \\
\verb|Transposition| & {1} \\
\verb|Text Reversal| & {2} \\ 
\verb|Character Shift| & {3} \\ 
\verb|Word Reversal| & {4} \\
\verb|Unencrypted| & {5}  \\ \hline
\end{tabular}
\caption{Label mapping}
\label{tab:EncodedLabels}
\end{table}

\begin{figure}[ht]
\centering
\includegraphics[width=\columnwidth]{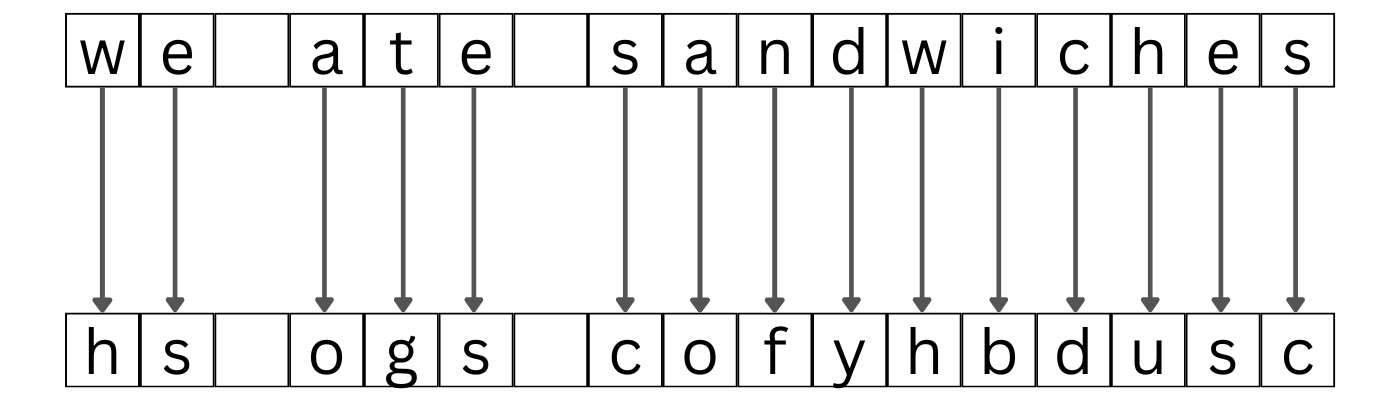}
\caption{Substitution Cipher}
\label{fig:SubstitutionCipher}
\end{figure}

\paragraph{Substitution Cipher:}
The substitution cipher replaces each letter in the alphabet with a new letter. We also added a derangement condition, which means that letters can not be replaced with themselves. This ensures that the generated substitution cipher is distinct from the unencrypted text. In Figure~\ref{fig:SubstitutionCipher}, we show a possible letter mapping of the substitution cipher. See figure~\ref{fig:SubstitutionCipher}.

\begin{figure}[ht]
\centering
\includegraphics[width=\columnwidth]{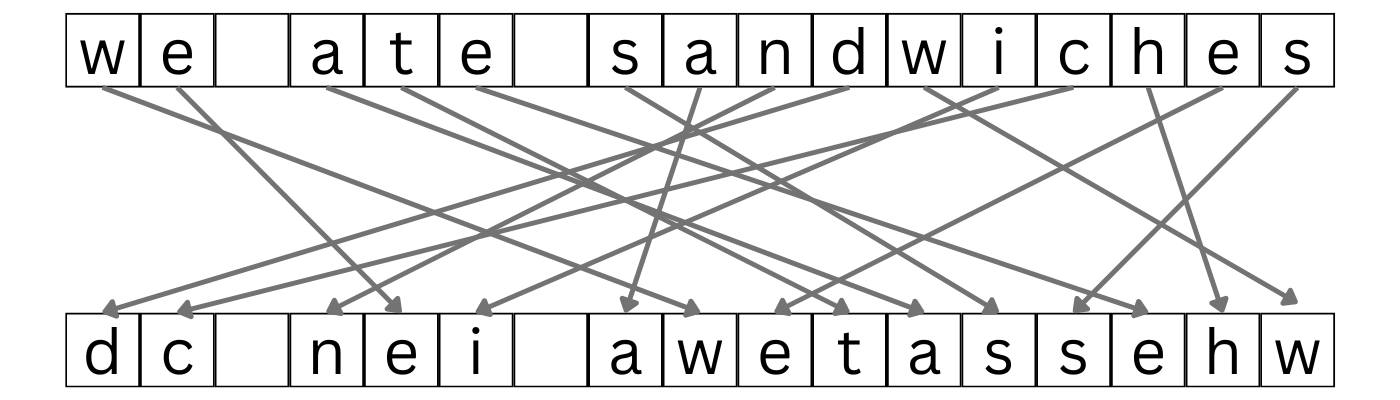}
\caption{Transposition Cipher}
\label{fig:TranspositionCipher}
\end{figure}

\paragraph{Transposition Cipher:}
The transposition cipher changes the position of each letter in the sentence. We add the condition to keep the whitespace characters in their original space. This method often results in letters mapping to themselves, but we add a condition to redo the transposition if the encrypted text matches the unencrypted text. See figure~\ref{fig:TranspositionCipher}.

\begin{figure}[ht]
\centering
\includegraphics[width=\columnwidth]{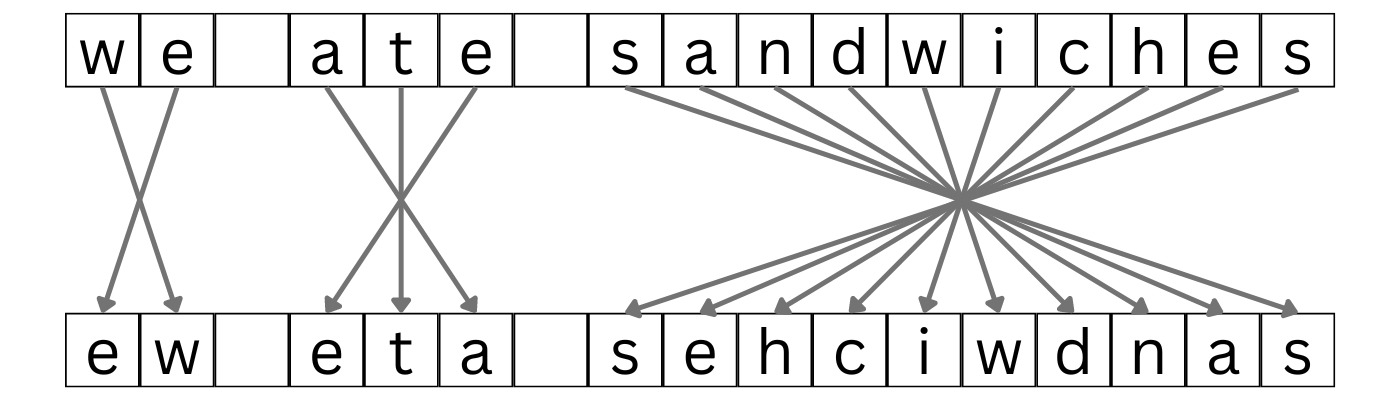}
\caption{Word Reversal}
\label{fig:WordReversal}
\end{figure}

\paragraph{Word Reversal:}
Word reversal is when every word in a sentence is reversed. The position of each whitespace character remains unchanged. See figure~\ref{fig:WordReversal}.

\begin{figure}[ht]
\centering
\includegraphics[width=\columnwidth]{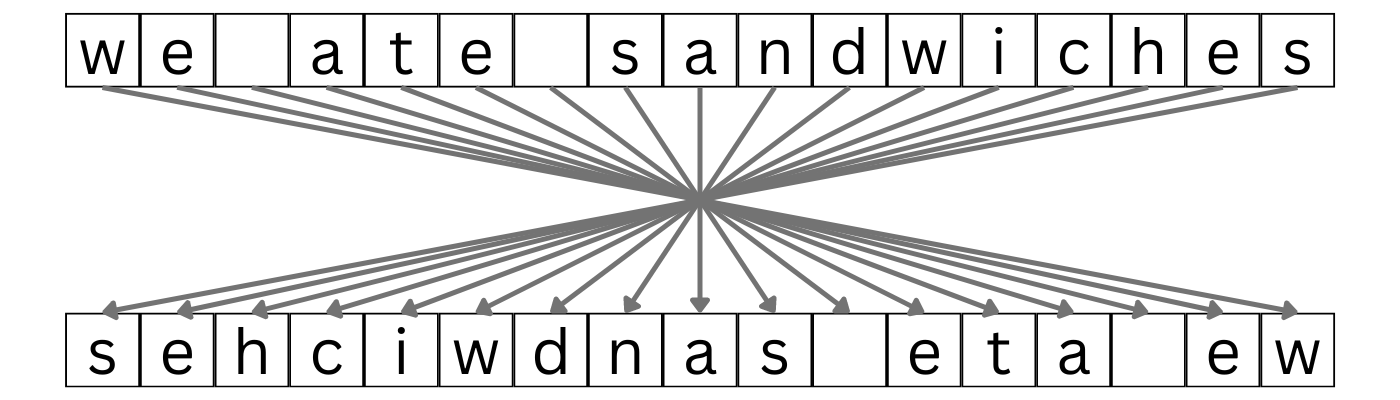}
\caption{Text Reversal}
\label{fig:TextReversal}
\end{figure}

\paragraph{Text Reversal:}
Text reversal is when every character in a sentence is reversed. This includes whitespace characters. See figure~\ref{fig:TextReversal}.

\begin{figure}[ht]
\centering
\includegraphics[width=\columnwidth]{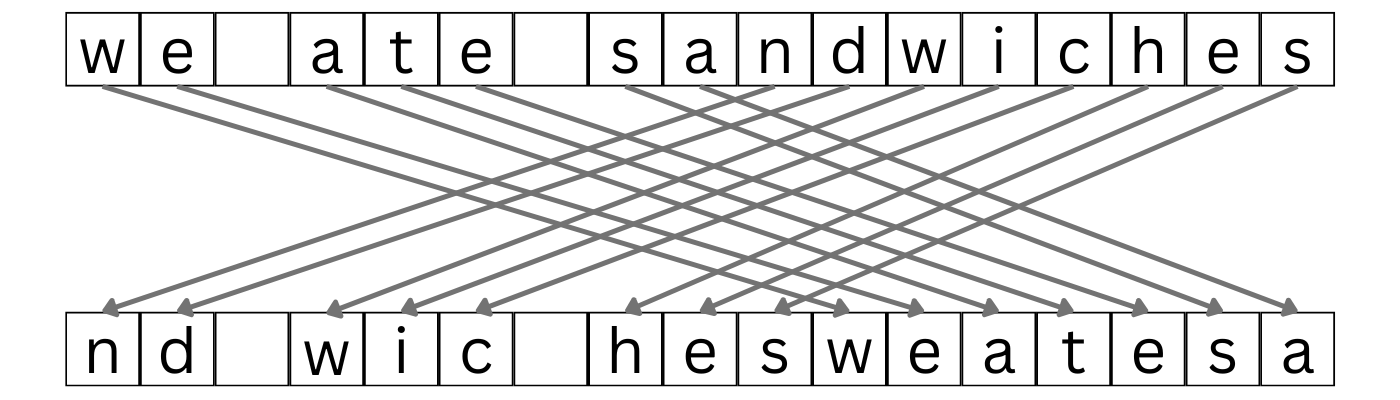}
\caption{Character shift}
\label{fig:CharacterShift}
\end{figure}

\paragraph{Character shift:}
Character shift is when all non whitespace characters are shifted in the sentence. See figure~\ref{fig:CharacterShift}.

\subsection{Models/Tokenizers}

The models tested were GRUs, LSTMs, and BERT. The GRU and LSTM models were implemented in Keras \citep{chollet2015keras}, and the BERT \citep{BERT} model was implemented in Pytorch \citep{pytorch}. The embedding dimension for each model was 300, except for BERT which had an embedding dimension of 768. At the end of model training, we would restore the model weights that achieved the highest accuracy on the validation dataset. 

The tokenization strategies tested were Byte-Pair encoding, WordPiece, character-level, and word-level tokenization. These were built using Python \citep{van1995python} and the tokenizers module from HuggingFace \citep{huggingface}.

We also tested pre-trained GloVe embeddings \citep{pennington-etal-2014-glove} as a baseline model. We did this because GloVe is a state-of-the-art static word embedding architecture used across many NLP tasks. The pre-trained GloVe embeddings were trained on the Wikipedia 2014 dataset and Gigaword corpora \citep{gigaword}. We also trained custom glove embedding weights using our corpora by modifying the code from the GloVe GitHub repository.

For the most part, the code for this project was implemented from scratch. To implement the frozen GloVe embeddings and the BERT training script, we used existing code. In these cases, we provide references to source code directly in the CipherSniffer code. Finally, all the code for this paper is open source and accessible in the CipherSniffer repository. We provide an example training notebook for those who want to reproduce the results from the paper. We also provide all the trained models and tokenizers for those who do not want to train them from scratch.

\subsection{General Architecture}

\begin{figure}[ht]
\centering
\includegraphics[width=\columnwidth]{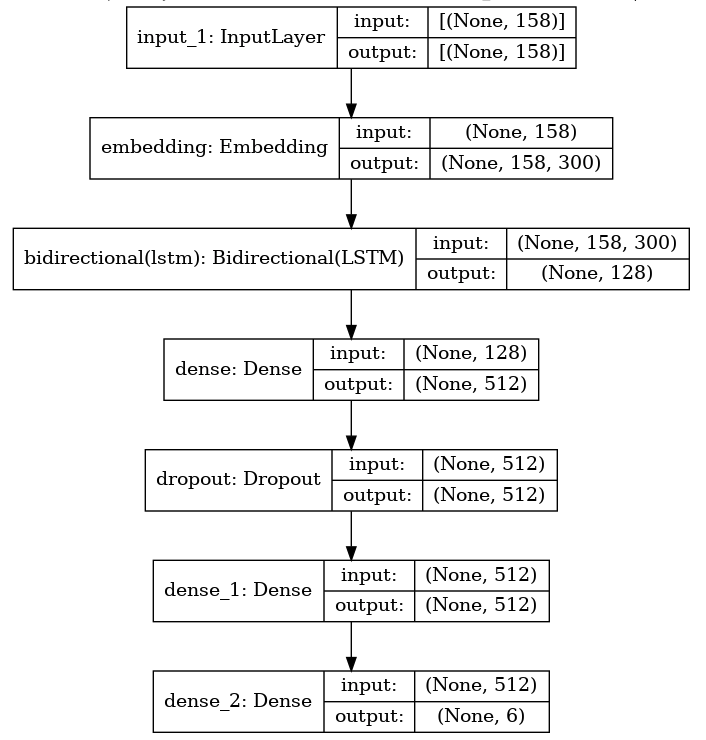}
\caption{LSTM Architecture}
\label{fig:ModelArchitecture}
\end{figure}

Each model takes an input of raw text and predicts an output class. Moreover, each model outputs a 6-dimensional vector scaled by the softmax function. Each scalar value is the predicted probability of the input text belonging to each class. We added a feed-forward network on top of each model to improve the accuracy and reduce the likelihood of overfitting. The feed-forward network structure is identical for each model. This network consists of two 512-dimensional linear layers with the ReLU activation function and 50\% dropout. See figure~\ref{fig:ModelArchitecture} for the LSTM model architecture.

\section{Experimental setup}

\subsection{Dataset}

The dataset that we used for this task was the Gigaword dataset \citep{gigaword}. This open-source dataset contains article summaries and headlines for 3,993,608 news articles. In our experiments, we only used the summaries from this dataset. We cleaned the data by lowercasing the text and removing words that contained non-ASCII-lowercase characters. This made the average sentence length 177 characters and reduced the character vocabulary from 128 to 27. 

Next, we created a subset of 55,000 data points for model training. We further split this set into 50,000 for training, 2,500 for validation, and 2,500 for an unseen test set. The remaining 3,943,608 article summaries were used as a corpus for the tokenization models. Due to the computational demand of training non-parallelizable RNN and LSTM models, we did not make the classifier training, validation, and test set any larger. See table~\ref{tab:CipherData} for more information on the dataset.

\begin{table}
\centering
\begin{tabular}{lcc}
\hline
\textbf{Data} & \textbf{Rows} & \textbf{Size (MB)} \\
\hline
Tokenizer & 3,943,608 & 660.8 \\
Train & 50,000 & 8.6 \\
Valid & 2,500 & 0.4 \\
Test & 2,500 & 0.4 \\
\hline
\end{tabular}
\caption{CipherData}
\label{tab:CipherData}
\end{table}

\subsection{Metrics}

During training, we optimized model parameters using the cross-entropy loss. The cross-entropy loss is the negative sum of each label denoted as “ti” multiplied by the log of the predicted probability of each class. Each “ti” value in the summation is zero, except for the true label which has a value of one. This is because there is only one true label for each data point. The equation for the cross-entropy loss is shown below. \\

\centerline{CE Loss $ = -\sum_{c=1}^My_{o,c}\log(p_{o,c}) $} ~\\
During model training, we monitored the training and validation accuracy at the end of each epoch. If the validation accuracy of the model was better than all previous epochs, we saved the model weights. This was done to ensure that we were not overfitting to the training data and that we had a highly accurate model. After the model finished training, we restored the saved model weights and performed a final evaluation on a hidden test set. \\

\centerline{Accuracy $ = \frac{TP+TN}{TP+TN+FP+FN} $}
\section{Results}
In table~\ref{tab:ScoreTable} we show the accuracy of each model. We trained each model for 10 epochs, with a learning rate of 0.001, a batch size of 1024, and an ADAM optimizer. Each model was trained on a single NVIDIA Tesla P100 GPU. Due to the number of model-tokenizer combinations, we did not perform any extensive hyperparameter tuning.

The BERT model with a wordpiece tokenizer achieved the highest overall test accuracy of 99.00\%. It correctly classified all data points in the substitution, transposition, character shift, and unencrypted text classes. It scored 99.04\% on the word reversal class and 90.65\% on the text reversal class. 

The second best model was a GRU with GloVe embeddings trained on our corpora. This model achieved an overall test accuracy of 97.00\%. It achieved the highest accuracy on the text reversal class with an accuracy of 95.20\%, and it also matched BERT on word reversals with a score of 99.04\%.

\subsection{Tokenizer Accuracy}

On average, the best tokenization strategy was the subword-level tokenizer. Models that used subword tokenizers scored an average overall accuracy of 96.13\%. We believe that this was due to their ability to capture common words and handle out-of-bounds vocabulary at the same time.

The word-level tokenizers were more unpredictable, scoring an average accuracy of 92.31\%. On the low end, the models that used frozen GloVe embeddings performed very poorly. They struggled to classify the substitution and transposition ciphers, scoring roughly 70-80\% accuracy on each class. This was likely due to the high number of out-of-bounds vocabulary created from applying ciphers to the text. On the high hand, the trained GloVe embeddings on our corpora of cipher text performed on par with the subword-level tokenizers.

The character-level tokenizers scored an average accuracy of 89.40\% and the results were quite interesting. They struggled to classify word reversals and text reversals with accuracy scores in the range of 58-69\% and 79-86\% respectively. Further, all the character-level tokenization models scored higher than the word-level tokenization models on the transposition and substitution classes. The results from the character-level tokenizer were more accurate than we expected, as there were only 28 possible embeddings in this model.

\subsection{Other}

All the models we tested had no problem classifying original text and character shifts. The classes that were difficult for all the models were the word reversal and text reversal classes. This was likely due to the similar character order of the ciphered text, but it is not entirely evident why the models struggled here. For more comprehensive model results, see table~\ref{tab:ScoreTable}.

We also recognize the difference in tunable parameters for each model. Moreover, the BERT uncased model has ~109 million tunable parameters compared with the GRU and LSTM models which have ~9.2 and ~9.3 million tunable parameters. These models have only 8\% of the tunable parameters of the BERT model, which could explain the difference in performance.

\section{Future Work}

It would be interesting to identify the root cause of some notable results in this paper. This includes the difficulty of classifying word reversals and character reversals and the high accuracy of a GRU model with GloVe Embeddings.

Second, human codebreakers typically use whitespace characters as cipher hints. This leads us to believe that a sentencepiece tokenizer \citep{SentencePiece} may outperform the BPE and wordpiece models. This tokenizer treats the input text as a sequence of Unicode characters rather than splitting on whitespace, and may provide more cipher hints to the model.

\begin{table*}[ht]
\centering
\begin{tabular}{l|lllllll|l}
\hline
\textbf{Model} & \textbf{Level} & \textbf{Subs} & \textbf{Trans} & \textbf{T-Rev} & \textbf{Chr-S} & \textbf{W-Rev} & \textbf{Original} & \textbf{Acc} \\
\hline
BERT & S & 1 & 1 & 0.9065 & 1 & 0.9904 & 1 & 0.9900 \\
GRU GLOVE & W & 0.9543 & 0.9448 & 0.9520 & 0.9784 & 0.9904 & 1 & 0.9700 \\
LSTM BPE & S & 1 & 0.9880 & 0.9041 & 0.9880 & 0.8681 & 0.9976 & 0.9576 \\
LSTM WP (pre) & S & 1 & 1 & 0.8609 & 0.9952 & 0.8897 & 0.9952 & 0.9568 \\
GRU WP (pre) & S & 0.9976 & 1 & 0.8129 & 0.9928 & 0.9400 & 0.9976 & 0.9568 \\
GRU WP & S & 1 & 0.9976 & 0.8609 & 0.9976 & 0.8825 & 1 & 0.9564 \\
GRU BPE & S & 1 & 1 & 0.8801 & 0.9880 & 0.8705 & 0.9976 & 0.9560 \\
LSTM WP & S & 1 & 0.9832 & 0.8321 & 0.9952 & 0.9257 & 0.9976 & 0.9556 \\
LSTM GLOVE & W & 0.9255 & 0.9544 & 0.9257 & 0.9832 & 0.9353 & 0.9976 & 0.9536 \\
LSTM Character Level & C & 0.9952 & 0.9976 & 0.7938 & 0.9544 & 0.6930 & 0.9952 & 0.9048 \\
GRU GLOVE (pre) & W & 0.7548 & 0.8106 & 0.8849 & 0.9065 & 0.9664 & 1 & 0.8872 \\
GRU Character Level & C & 0.9880 & 0.9736 & 0.8609 & 0.9329 & 0.5803 & 0.9639 & 0.8832 \\
LSTM GLOVE (pre) & W & 0.8005 & 0.7146 & 0.9353 & 0.9233 & 0.9185 & 0.9976 & 0.8816 \\
\hline
Avg Accuracy &  & 0.9551 & 0.9511 & 0.8777 & 0.9720 & 0.8808 & 0.9954 & 
\end{tabular}
\caption{Model Accuracy \\\hspace{\textwidth}}
\label{tab:ScoreTable}
\parbox{\textwidth}{\small{\textbf{Table Abbreviations:} S(Subword), W(Word), C(Character), Subs(Substitution), Trans(Transposition), T-Rev(Text Reversal), Chr-S(Character Shift), W-Rev(Word Reversal)}}
\end{table*}

\label{sec:bibtex}






\bibliography{custom}

\begin{thebibliography}{13}
\expandafter\ifx\csname natexlab\endcsname\relax\def\natexlab#1{#1}\fi

\bibitem[{Berg-Kirkpatrick and
  Klein(2013)}]{berg-kirkpatrick-klein-2013-decipherment}
Taylor Berg-Kirkpatrick and Dan Klein. 2013.
\newblock \href {https://aclanthology.org/D13-1087} {Decipherment with a
  million random restarts}.
\newblock In \emph{Proceedings of the 2013 Conference on Empirical Methods in
  Natural Language Processing}, pages 874--878, Seattle, Washington, USA.
  Association for Computational Linguistics.

\bibitem[{Chollet et~al.(2015)}]{chollet2015keras}
Francois Chollet et~al. 2015.
\newblock \href {https://github.com/fchollet/keras} {Keras}.

\bibitem[{Devlin et~al.(2018)Devlin, Chang, Lee, and Toutanova}]{BERT}
Jacob Devlin, Ming{-}Wei Chang, Kenton Lee, and Kristina Toutanova. 2018.
\newblock \href {http://arxiv.org/abs/1810.04805} {{BERT:} pre-training of deep
  bidirectional transformers for language understanding}.
\newblock \emph{CoRR}, abs/1810.04805.

\bibitem[{Graff et~al.(2003)Graff, Kong, Chen, and Maeda}]{gigaword}
David Graff, Junbo Kong, Ke~Chen, and Kazuaki Maeda. 2003.
\newblock English gigaword.
\newblock \emph{Linguistic Data Consortium, Philadelphia}, 4(1):34.

\bibitem[{Kambhatla et~al.(2018)Kambhatla, Mansouri~Bigvand, and
  Sarkar}]{kambhatla-etal-2018-decipherment}
Nishant Kambhatla, Anahita Mansouri~Bigvand, and Anoop Sarkar. 2018.
\newblock \href {https://doi.org/10.18653/v1/D18-1102} {Decipherment of
  substitution ciphers with neural language models}.
\newblock In \emph{Proceedings of the 2018 Conference on Empirical Methods in
  Natural Language Processing}, pages 869--874, Brussels, Belgium. Association
  for Computational Linguistics.

\bibitem[{Krishna et~al.(2019)Krishna, Stamp, Austin, and
  Troia}]{krishna-etal-2018-classify}
Krishna, Stamp, Austin, and Troia. 2019.
\newblock \href {https://doi.org/10.31979/etd.xkgs-5gy6} {Classifying classic
  ciphers using machine learning}.

\bibitem[{Kudo and Richardson(2018)}]{SentencePiece}
Taku Kudo and John Richardson. 2018.
\newblock \href {http://arxiv.org/abs/1808.06226} {Sentencepiece: {A} simple
  and language independent subword tokenizer and detokenizer for neural text
  processing}.
\newblock \emph{CoRR}, abs/1808.06226.

\bibitem[{Paszke et~al.(2019)Paszke, Gross, Massa, Lerer, Bradbury, Chanan,
  Killeen, Lin, Gimelshein, Antiga, Desmaison, Kopf, Yang, DeVito, Raison,
  Tejani, Chilamkurthy, Steiner, Fang, Bai, and Chintala}]{pytorch}
Adam Paszke, Sam Gross, Francisco Massa, Adam Lerer, James Bradbury, Gregory
  Chanan, Trevor Killeen, Zeming Lin, Natalia Gimelshein, Luca Antiga, Alban
  Desmaison, Andreas Kopf, Edward Yang, Zachary DeVito, Martin Raison, Alykhan
  Tejani, Sasank Chilamkurthy, Benoit Steiner, Lu~Fang, Junjie Bai, and Soumith
  Chintala. 2019.
\newblock \href
  {http://papers.neurips.cc/paper/9015-pytorch-an-imperative-style-high-performance-deep-learning-library.pdf}
  {Pytorch: An imperative style, high-performance deep learning library}.
\newblock In \emph{Advances in Neural Information Processing Systems 32}, pages
  8024--8035. Curran Associates, Inc.

\bibitem[{Pennington et~al.(2014)Pennington, Socher, and
  Manning}]{pennington-etal-2014-glove}
Jeffrey Pennington, Richard Socher, and Christopher Manning. 2014.
\newblock \href {https://doi.org/10.3115/v1/D14-1162} {{G}lo{V}e: Global
  vectors for word representation}.
\newblock In \emph{Proceedings of the 2014 Conference on Empirical Methods in
  Natural Language Processing ({EMNLP})}, pages 1532--1543, Doha, Qatar.
  Association for Computational Linguistics.

\bibitem[{Rayner et~al.(2006)Rayner, White, Johnson, and
  Liversedge}]{jumbled-letters}
Keith Rayner, Sarah White, Rebecca Johnson, and Simon Liversedge. 2006.
\newblock \href {https://doi.org/10.1111/j.1467-9280.2006.01684.x} {Raeding
  wrods with jubmled lettres: There is a cost}.
\newblock \emph{Psychological science}, 17:192--3.

\bibitem[{Sakaguchi et~al.(2016)Sakaguchi, Duh, Post, and
  Van~Durme}]{robust-word-recognition}
Keisuke Sakaguchi, Kevin Duh, Matt Post, and Benjamin Van~Durme. 2016.
\newblock \href {https://doi.org/10.48550/ARXIV.1608.02214} {Robsut wrod
  reocginiton via semi-character recurrent neural network}.

\bibitem[{Van~Rossum and Drake~Jr(1995)}]{van1995python}
Guido Van~Rossum and Fred~L Drake~Jr. 1995.
\newblock \emph{Python reference manual}.
\newblock Centrum voor Wiskunde en Informatica Amsterdam.

\bibitem[{Wolf et~al.(2020)Wolf, Debut, Sanh, Chaumond, Delangue, Moi, Cistac,
  Rault, Louf, Funtowicz, Davison, Shleifer, von Platen, Ma, Jernite, Plu, Xu,
  Le~Scao, Gugger, Drame, Lhoest, and Rush}]{huggingface}
Thomas Wolf, Lysandre Debut, Victor Sanh, Julien Chaumond, Clement Delangue,
  Anthony Moi, Pierric Cistac, Tim Rault, Remi Louf, Morgan Funtowicz, Joe
  Davison, Sam Shleifer, Patrick von Platen, Clara Ma, Yacine Jernite, Julien
  Plu, Canwen Xu, Teven Le~Scao, Sylvain Gugger, Mariama Drame, Quentin Lhoest,
  and Alexander Rush. 2020.
\newblock \href {https://doi.org/10.18653/v1/2020.emnlp-demos.6} {Transformers:
  State-of-the-art natural language processing}.
\newblock In \emph{Proceedings of the 2020 Conference on Empirical Methods in
  Natural Language Processing: System Demonstrations}, pages 38--45, Online.
  Association for Computational Linguistics.

\end{thebibliography}
\bibliographystyle{acl_natbib}

\appendix



\end{document}